%% file: main.tex
\documentclass[11pt, a4paper, logo, copyright, nonumbering]{map}
\input{resources/packages}

\definecolor{hidden-draw}{RGB}{20,68,106}
\definecolor{hidden-pink}{RGB}{255,245,247}
\usepackage[edges]{forest}

\usepackage{bm}
\usepackage{tikz}
\usetikzlibrary{arrows.meta,positioning}
\usepackage{natbib}
\usepackage{CJKutf8}
\usepackage{xargs}
\usepackage{todonotes}
\usepackage{multirow}
\usepackage{cleveref}
\usepackage{amsmath}
\usepackage[cal=cm,bb=ams]{mathalfa}
\usepackage{dsfont}
\usepackage{subcaption}
\usepackage{url}
\usepackage{svg}
\usepackage{makecell}
\usepackage[table]{xcolor}
\usepackage{fontawesome}
\usepackage{xcolor}
\usepackage{float}

\usepackage[utf8]{inputenc}
\usepackage{booktabs}
\usepackage{graphicx}
\usepackage{array}
\usepackage{tabularx}
\usepackage{xcolor,colortbl}  
\definecolor{boxcolor}{HTML}{d92523} 
\definecolor{bulbcolor}{HTML}{e3b87f} 
\usepackage{url}
\usepackage{ragged2e}   
\usepackage{makecell} 

\usepackage{hyperref}       
\usepackage{url}            
\usepackage{booktabs}       
\usepackage{amsfonts}       
\usepackage{nicefrac}       
\usepackage{microtype}      
\usepackage{xcolor}         
\usepackage{graphicx}
\usepackage{longtable}
\usepackage{array}
\usepackage{pbox}
\usepackage{amsmath}
\usepackage{amssymb}
\usepackage{tabularx}
\usepackage{multirow}
\usepackage{wrapfig}
\usepackage{pifont}
\usepackage{algorithm}
\usepackage{algorithmic}
\usepackage{lipsum}
\usepackage{subcaption}
\usepackage{enumitem}
\usepackage{tikz}
\usepackage{xstring}

\usepackage{color-edits}

\usepackage{booktabs}
\usepackage{multirow}
\usepackage[table]{xcolor}
\usepackage{tabularx}

\usepackage{parskip} 
\usepackage{threeparttable} 

\newcommand{\modelname}{IQuest-Q1} 
\definecolor{rliableolive}{HTML}{BBCC33}
\definecolor{rliableblue}{HTML}{77AADD}
\definecolor{rliablered}{HTML}{f63c44}

\definecolor{rliableolive}{HTML}{BBCC33}
\definecolor{rliableblue}{HTML}{77AADD}
\definecolor{rliablered}{HTML}{f63c44}

\usepackage[most,skins,theorems]{tcolorbox}
\tcbuselibrary{skins,breakable,fitting,hooks}  
\tcbset{
  aibox/.style={
    width=\linewidth,
    top=7pt,
    bottom=2pt,
    colback=rliablered!18!white,
    colframe=black,
    colbacktitle=black,
    enhanced,
    center,
    attach boxed title to top left={yshift=-0.1in,xshift=0.15in},
    boxed title style={boxrule=0pt,colframe=white,},
  }
}
\tcbset{
  aibox2/.style={
    width=\linewidth,
    top=7pt,
    bottom=2pt,
    colback=green!18!white,
    colframe=black,
    colbacktitle=black,
    enhanced,
    center,
    attach boxed title to top left={yshift=-0.1in,xshift=0.15in},
    boxed title style={boxrule=0pt,colframe=white,},
  }
}
\tcbset{
  rsiquote/.style={
    enhanced,
    breakable,
    width=\linewidth,
    colback=IQuestAbstractPale,
    colframe=IQuestAbstractPale,
    coltext=IQuestAbstractInk,
    boxrule=0pt,
    borderline west={1.6mm}{0pt}{IQuestAbstractSky},
    arc=3mm,
    auto outer arc,
    left=4mm,
    right=4mm,
    top=3mm,
    bottom=3mm,
    before skip=0.8\baselineskip,
    after skip=0.8\baselineskip
  }
}
\newtcolorbox{RSIQuote}{rsiquote}
\newtcolorbox{AIbox}[2][]{aibox,title=#2,#1}
\newtcolorbox{AIbox2}[2][]{aibox2,title=#2,#1}

\hypersetup{
    colorlinks=true,            
    linkcolor=blue,             
    filecolor=magenta,          
    urlcolor=cyan,              
    citecolor=purple,             
    pdftitle={Overleaf Example},
    pdfpagemode=FullScreen,
}

\definecolor{iquestblue}{HTML}{173C7F}
\definecolor{iquestazure}{HTML}{528FCC}

\renewcommand{\today}{}
\newcommandx{\info}[2][1=]{\todo[linecolor=red,backgroundcolor=red!25,bordercolor=red,#1]{#2}}

\title{
\vspace{-0.2in}
\centering \fontsize{15pt}{16pt}\selectfont
\modelname{}: Advancing Foundation Capabilities \\ for Agentic CLI Systems 

\vspace{-0.2in}
}

\author{
IQuest Research
\vspace{-5pt}
}

\input{sections/abstract}

\begin{document}

\maketitle

\let\oldthefootnote\thefootnote
\begin{figure*}[h!]
    \centering
    \includegraphics[width=0.99\textwidth]{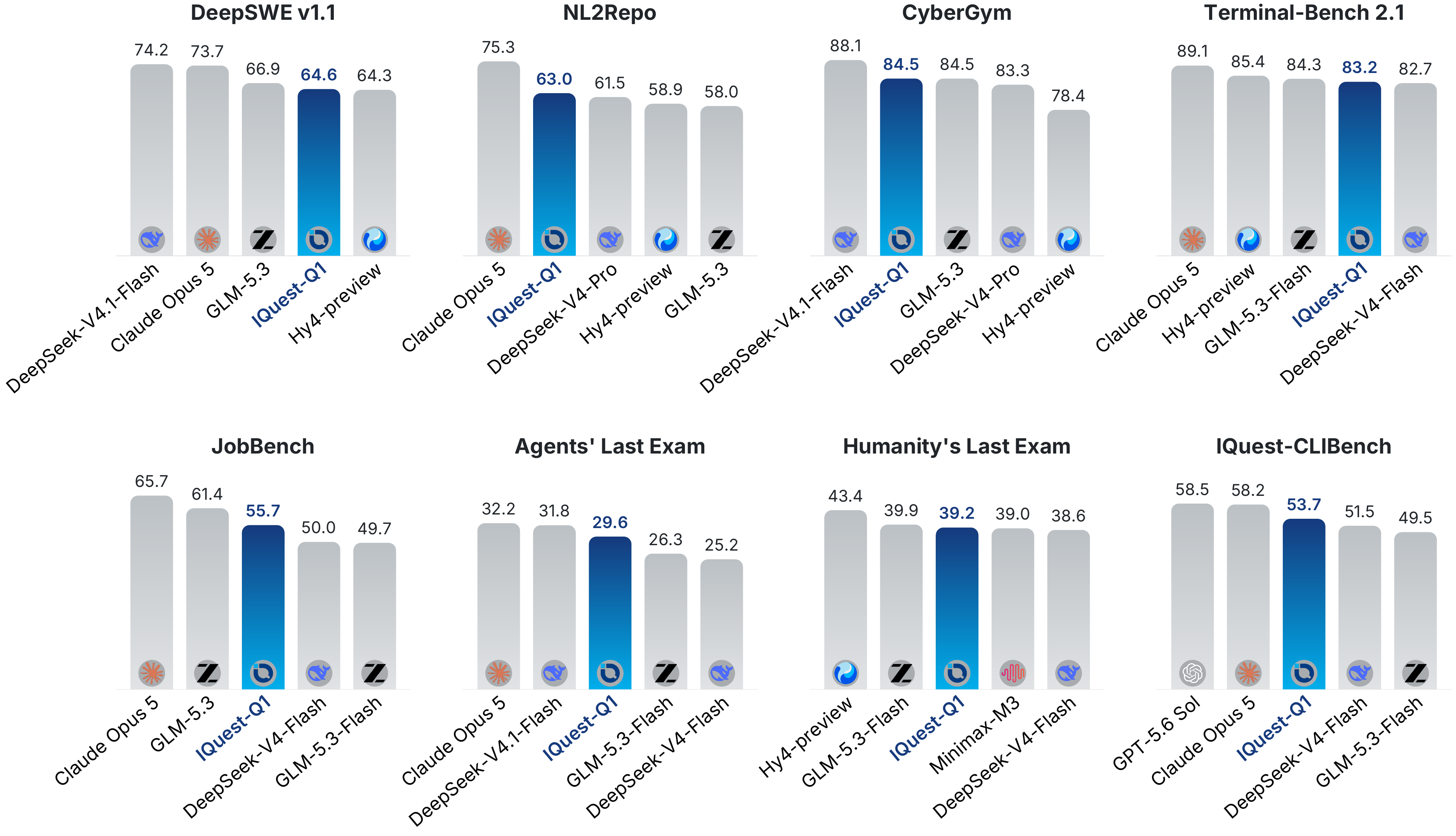}
    \caption{\modelname{} performance across benchmarks. DeepSeek-V4-Flash/Pro refer to the official 0731/0813 releases, respectively. Humanity's Last Exam is reported without tools. IQuest-CLIBench is our proprietary built-in benchmark for evaluating human-agent collaboration and user experience (see Section~\ref{sec:evaluation} for details).}
    \label{fig:tease_perf}
\end{figure*}

\newpage

\newpage

\input{sections/intro}
\input{sections/model_architecture}

\input{sections/pre_training}
\input{sections/post_training}

\input{sections/evaluation}

\input{sections/case_studies}
\input{sections/limitations}
\input{sections/conclusion}

\clearpage
\newpage

\input{sections/author_list}

\bibliography{ref}

\input{sections/appendix}

\end{document}

%% file: resources/packages.tex
\usepackage{etex}
\usepackage{CJKutf8}
\usepackage{xargs}  
\usepackage{todonotes}
\usepackage{amsmath}
\usepackage{dsfont}

\usepackage{longtable}

\usepackage{svg}
\usepackage{fontawesome}
\usepackage{array}
\usepackage{tabularx}
\usepackage{latexsym}
\usepackage{graphicx}
\usepackage[utf8]{inputenc} 
\usepackage[utf8]{inputenc} 
\usepackage{amssymb} 
\usepackage{url}            
\usepackage{amsfonts}       
\usepackage{nicefrac}       
\usepackage{microtype}      
\usepackage{xspace}

\usepackage{listings}
\usepackage{makecell}
\usepackage{inconsolata}
\usepackage{multicol}
\usepackage{amstext}

\definecolor{darkblue}{RGB}{84, 112, 198}

\definecolor{lightblue}{rgb}{0.85, 0.95, 1.0}    
\definecolor{lightgreen}{rgb}{0.90, 1.0, 0.90}    
\definecolor{lightorange}{rgb}{1.0, 0.95, 0.85}   
\definecolor{lightpurple}{rgb}{0.95, 0.90, 1.0}   
\definecolor{lightgray}{rgb}{0.97, 0.97, 0.97}    

\usepackage{tikz}
\usetikzlibrary{calc}
\definecolor{battery-empty}{rgb}{0.9, 0.9, 0.9}
\newcommand{\difficultybar}[1]{%
  \begin{tikzpicture}[baseline, scale=0.5, every node/.style={scale=0.8}]
    \foreach \i in {1,2,3,4,5} {
      \ifnum\i>#1
        \draw[fill=battery-empty] (\i*0.5-0.5, 0) rectangle (\i*0.5, 0.25);
      \else
        \pgfmathsetmacro{\colorlevel}{80 - 12*(\i)} 
        \edef\x{\noexpand\draw[fill=blue!\colorlevel!white, opacity=0.9] (\i*0.5-0.5, 0) rectangle (\i*0.5, 0.25);}
        \x
        \draw[blue!50!black] (\i*0.5-0.5, 0) rectangle (\i*0.5, 0.25);
      \fi
    }
    \fill[battery-empty!70] (2.5, 0.08) rectangle (2.6, 0.17);
    \draw[battery-empty!70!black] (2.5, 0.08) rectangle (2.6, 0.17);
  \end{tikzpicture}%
}
\renewcommand{\arraystretch}{0.96}

%% file: sections/abstract.tex
\begin{abstract}
We present \modelname{}, a sparse Mixture-of-Experts (MoE) agentic foundation model with 320B total parameters (15B activated).
\modelname{} is developed through a three-stage process:
(1) \emph{stable and efficient pre-training}, combining Stable SplitNorm, Muon, and $\mu$P with a two-stage curriculum whose second stage increases the coding-data mixture, preserving a wide stability margin between safe and optimal learning rates;
(2) \emph{fine-grained mid-training}, expanding the data mixture across domains and formats and progressively extending the context;
and
(3) \emph{multi-scaffold with multi-version post-training}, combining SFT, domain-specialized RL, and MOPD across various scaffolds to integrate complementary training signals and strengthen agentic coding capabilities.
Furthermore, model merging combines different experts and checkpoints from different stages while mitigating conflicts among their parameter updates.
The resulting \modelname{} demonstrates capabilities in both coding and general tasks.
We release \modelname{} as an open-source model to support community research and development.
\end{abstract}

%% file: sections/intro.tex
\section{Introduction}
\label{sec:introduction}



LLM-based agents allow language models~\cite{openai2026gpt56systemcard,anthropic2026claudeopus5,deepseekai2026deepseekv4,zeng2026glm,kimiteam2026kimik3openfrontier,yang2026iquest,yang2026incoder32bthinkingindustrialcodeworld} to perform tasks through iterative reasoning, action, and interaction with external environments, rather than merely generating responses~\cite{openai2025gpt5codex,anthropic2025claudecode,yang2025code}. Among these environments, command-line interfaces (CLIs) have become one of the most general ways for language models to interact with external systems, use tools, access files, and execute code.
As a result, building more capable agentic models requires stronger CLI capabilities.


In this work, we release \modelname{}, an open-source agentic foundation model with broad CLI capabilities across a wide range of general and coding tasks, with strengths in software engineering, general-purpose task execution, long-horizon reasoning and execution, and cybersecurity. 
We also optimize \modelname{} for real-world CLI use, with reliable tool use and robust performance across different agent harnesses.

\begin{figure*}[t!]
    \centering
    \includegraphics[width=\textwidth]{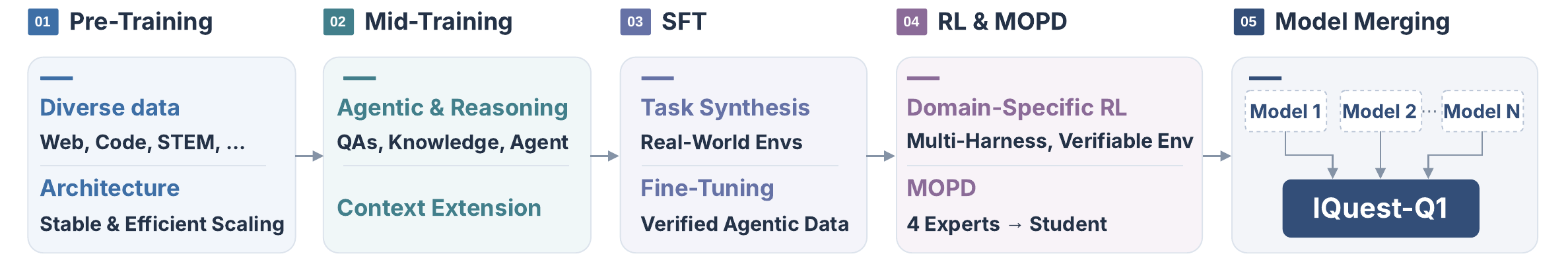}
    \caption{Training pipeline of \modelname{}.}
    \label{fig:pipeline}
\end{figure*}

\paragraph{Methods.} \modelname{} is trained through a multi-stage pipeline (see~\autoref{fig:pipeline}) to progressively develop general knowledge, long-context capabilities, and agentic CLI capabilities.
General pre-training combines Stable SplitNorm~\citep{zhuo2026hybridnorm,wechat_ai_team2026building}, Muon~\citep{liu2025muonscalable}, and $\mu$P  parameterization~\citep{yang2022tensorprogramsv} with a two-stage curriculum that increases the coding-data mixture in the second stage. This design improves scaling efficiency while maintaining a wide stability margin between safe and optimal learning rates. 
Mid-training then expands the data mixture across domains and formats while progressively extending the context.
Post-training (shown in~\autoref{fig:IQuest-Q1-Post-Training}) builds diverse CLI~\citep{openai2025gpt5codex, liumdeval} capabilities through synthesized tasks executed in production environments.
Specifically, SFT establishes basic CLI interaction, while RL enhances capabilities through task-level feedback and multi-teacher on-policy distillation (MOPD) consolidates complementary model capabilities from diverse frozen experts into a unified model. 
Model merging further combines complementary capabilities from expert models trained for different CLIs, while mitigating conflicts among their parameter updates.

\paragraph{Results.}
\autoref{fig:tease_perf} shows the results of \modelname{} on 8 agentic and reasoning benchmarks, highlighting three aspects of its capabilities:
(1) \emph{Advancing coding capability.} \modelname{} demonstrates coding capabilities across benchmarks, including DeepSWE v1.1, CyberGym, and NL2Repo.
(2) \emph{Broad general agentic and reasoning capability.} Beyond coding, \modelname{} demonstrates capabilities on general agentic and reasoning benchmarks, including JobBench, Agents' Last Exam, and Humanity's Last Exam.
(3) \emph{Optimized CLI interaction experience.} On our internal IQuest-CLIBench, \modelname{} manifests user-facing CLI interaction capabilities on interactive task.

Beyond benchmark performance, \modelname{} participates in an iterative model-development process built on our internal infrastructure, contributing to the development of its own successors under human supervision. 
This human-on-the-loop process provides a practical pathway towards recursive self-improvement, allowing model-generated improvements to accelerate subsequent model iterations.

Our contributions are summarized as follows:
\begin{itemize}

    \item \textbf{\modelname{}: an open-weight agentic model optimized for CLI capability.}
    \modelname{} is a sparse MoE model with 320B total parameters, of which 15B are activated per token, combining substantial general-purpose, coding-agent, and CLI interaction capabilities.
    We release its model weights to support further research and development in the community.
  
    \item \textbf{CLI-centric training pipeline.}
    We build the training pipeline around CLI-based agentic interaction. Pre-training progressively develops the capabilities required for downstream interaction through staged data mixtures. Post-training specializes these capabilities through task and environment synthesis, multi-scaffold RL, and MOPD to consolidate multiple models into a unified model.

    \item \textbf{Model development with human-on-the-loop RSI.}
    We incorporate \modelname{} into our iterative model-development process, allowing it to evaluate its capabilities and suggest potential improvements through a unified infrastructure interface under human supervision. 
    This process enables continuous model improvement while shortening the overall development cycle.
 
\end{itemize}

%% file: sections/model_architecture.tex
\section{Model Architecture}

\subsection{Overall Architecture}

\begin{figure}[!t]
    \centering
    \includegraphics[width=0.76\linewidth]{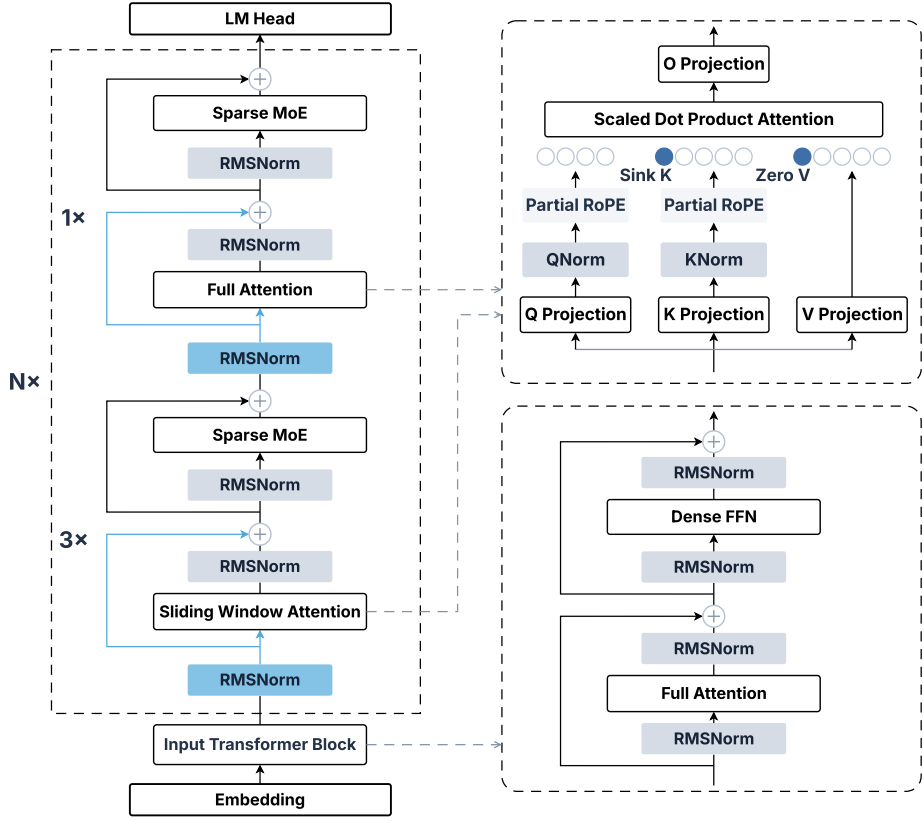}
    \caption{Overall architecture of \modelname{}.}
    \label{fig:overall-model-architecture}
\end{figure}

\modelname{} builds on the Transformer architecture~\citep{vaswani2017attention} with sparse Mixture-of-Experts (MoE) layers~\citep{fedus2022switchtransformers} and hybrid sliding-window attention (see \autoref{fig:overall-model-architecture}). The attention stack interleaves Grouped-Query Attention (GQA)~\citep{ainslie2023gqa} and Sliding Window Attention (SWA)~\citep{jiang2023mistral7b}. Within each attention module, a query-aware sink token~\citep{agarwal2025gpt} provides an adaptive destination for excess attention mass. Query--key normalization (QKNorm) stabilizes attention by controlling query and key scales~\citep{henry2020querykey}, and QKNorm Clip is additionally applied to prevent excessively large affine parameters. Partial RoPE applies rotary position embeddings~\citep{su2024roformer} to only a subset of the dimensions in each attention head. On the feed-forward side, the first Transformer layer replaces sparse MoE routing with a dense MLP to mitigate expert load imbalance in the earliest layer of the model.
To improve scaling effectiveness, \modelname{} adopts Stable SplitNorm to increase effective depth~\citep{csordas2025do,schwarzschild2022the}. For efficient and stable scaling, we combine the Muon optimizer~\citep{liu2025muonscalable} with $\mu$P parameterization~\citep{yang2022tensorprogramsv}, supporting efficient optimization and learning-rate transfer across model sizes.


\paragraph{Stable SplitNorm.}

Traditional Pre-Norm~\citep{xiong2020layer} may underutilize the depth of the model. As the unnormalized residual stream accumulates across layers, its variance can grow with depth, reducing the relative contribution of later residual updates~\citep{sun2026curse}. Motivated by these observations, \modelname{} adopts a hybrid normalization design introduced in prior work~\citep{zhuo2026hybridnorm,wechat_ai_team2026building}. We refer to this design as \emph{SplitNorm}, which splits the normalization strategy across Transformer sublayers: the residual stream entering the attention sublayer is normalized, whereas the residual stream entering the feed-forward network (FFN) is kept unnormalized. This design combines the stability benefits of Pre-Norm with the stronger depth utilization associated with Post-Norm, mitigating the underutilization of deeper layers in Pre-Norm architectures. Given an input $x$, the SplitNorm block used here is formulated using RMSNorm~\citep{zhang2019rmsnorm} as:
\begin{equation}
    z = \mathrm{RMSNorm}(x), \qquad h = z + \mathrm{Attention}(z), \qquad y = h + \mathrm{FFN}\bigl(\mathrm{RMSNorm}(h)\bigr).
\label{eq:SplitNorm}
\end{equation}
where $x$ denotes the input hidden state, $z$ denotes the RMSNorm-normalized input, $h$ is the intermediate representation after the attention block with a residual connection, and $y$ is the output after the FFN block with another residual connection.

This hybrid design combines normalized attention states with an identity residual path around the FFN. However, directly scaling this formulation to large models can lead to unstable optimization. To address this issue, \modelname{} further introduces two modifications. First, an RMSNorm is applied to the attention output to prevent activation explosion, such that the attention update in \autoref{eq:SplitNorm} becomes \(h = z + \mathrm{RMSNorm}(\mathrm{Attention}(z))\). Second, a SandwichNorm structure~\citep{ding2021cogview}, implemented with RMSNorm, is used for the first layer to stabilize the model input at the beginning of the residual stream. 

\paragraph{$\mu$P: Transfer Across Width and Depth.}

\modelname{} adopts $\mu$P parameterization~\citep{yang2022tensorprogramsv} for width-wise learning-rate transfer. Depth-wise transfer depends on additional parameterization choices, as studied in Depth-$\mu$P~\citep{yang2023tensorprogramsvi} and CompleteP~\citep{dey2025completep}. Under our adopted parameterization, empirical validation demonstrates hyper-parameter transfer across models spanning a $4\times$ range in width and a $4\times$ range in depth. Instead of tuning the learning rate directly at the target scale, we perform the search on a small proxy model with a hidden width of $896$ and $18$ layers. The learning rate selected on this proxy is then transferred zero-shot to the target model under the adopted parameterization, without additional learning-rate tuning at the target scale. We provide detailed $\mu$P scaling factors in \autoref{tab:mup_scaling_factors} in the appendix.

\paragraph{Recursive Multi-token Prediction.}

To improve backbone learning through additional future-token supervision, we incorporate two multi-token prediction (MTP)~\citep{gloeckle2024better,deepseekai2024deepseekv3technicalreport} layers from the start of pre-training. After completing the training pipeline, we adapt the first layer into a recursive drafter for speculative decoding and discard the second layer in a dedicated training stage. As in EAGLE-3~\citep{li2026eagle}, we extend the speculation horizon with a fixed parameter budget by reusing the same MTP layer across seven drafting steps~\citep{cai2025fastmtp,zeng2026glm}. In each drafting step, the output hidden state and generated token are fed back into the same MTP layer to produce subsequent predictions. We also replace its attention with SWA using a smaller window, which bounds the per-step attention cost as the context grows. Details can be found in \autoref{fig:mtp} in the appendix.

\subsection{Scaling Efficiency and Stability}

\paragraph{Scaling efficiency.}

\begin{figure*}[t]
    \centering
    \begin{subfigure}[t]{0.47\textwidth}
        \centering
        \includegraphics[width=\linewidth]{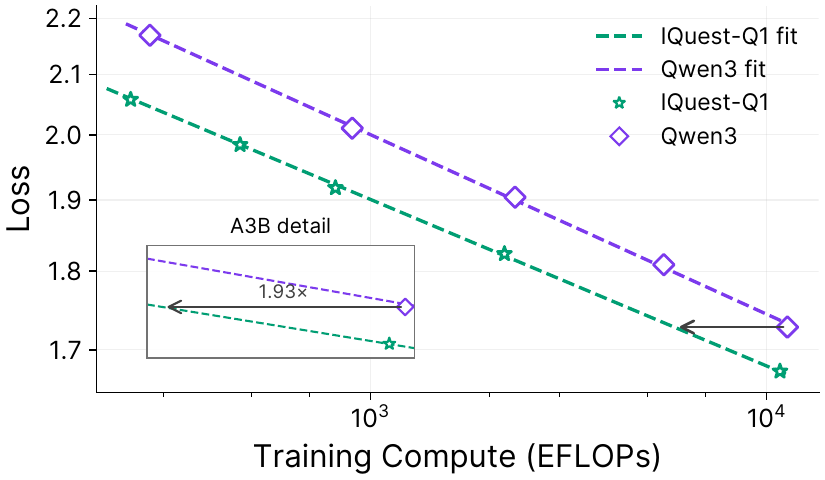}
        \caption{Scaling efficiency of \modelname{} and Qwen3-MoE using identical training data.}
        \label{fig:scaling_efficiency}
    \end{subfigure}
    \hfill
    \begin{subfigure}[t]{0.47\textwidth}
        \centering
        \includegraphics[width=\linewidth]{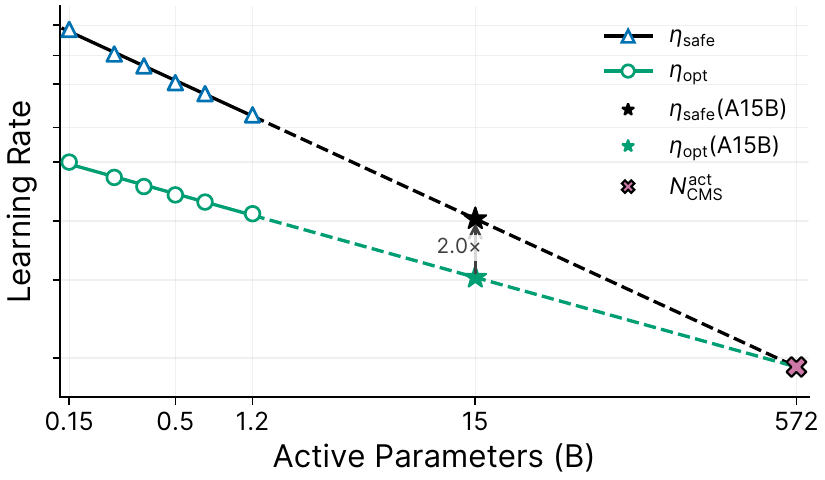}
        \caption{Maximum safe and optimal learning rates across activated-parameter scales.}
        \label{fig:max_safe_lr}
    \end{subfigure}
    \caption{Scaling efficiency and stability of \modelname{}.}
    \label{fig:scaling_stability_efficiency}
\end{figure*}

To evaluate scaling efficiency, we conduct a controlled comparison between \modelname{} and a representative Qwen3-MoE-style design~\citep{yang2025qwen3} following the Qwen3-30B-A3B configuration, using identical training data.
As shown in \autoref{fig:scaling_efficiency}, \modelname{} consistently achieves lower training cross-entropy loss at a given training FLOPs budget. Overall, \modelname{} exhibits a $1.93\times$ higher scaling efficiency than the baseline, requiring $1/1.93 \approx 52\%$ of the training compute to reach the same training loss. These results demonstrate the cumulative benefit of the proposed model design under controlled training conditions.

\paragraph{Scaling stability.} 
Inspired by the Critical Batch Size (CBS) perspective~\citep{mccandlish2018empirical,merrill2026critical}, we propose a complementary framework for characterizing the scaling limit imposed by training stability. CBS identifies the largest batch size that can be used without substantially degrading optimization efficiency; analogously, our framework aims to identify the largest model size that can be trained under a given architecture and optimization recipe without sacrificing either stability or optimization efficiency.

For a model of size $N$, we consider two learning-rate boundaries: the \emph{optimal learning rate} $\eta_{\mathrm{opt}}(N)$, which gives the most efficient optimization under a fixed training budget, and the \emph{safe learning rate} $\eta_{\mathrm{safe}}(N)$, defined as the largest learning rate that remains within the stable training regime. We provide the definition and evaluation criteria for stable training in Appendix~\ref{app:pre-training:safe-lr}. The available stability headroom can be quantified as:
\begin{equation}
H_N
=
\log
\frac{\eta_{\mathrm{safe}}(N)}
{\eta_{\mathrm{opt}}(N)}.
\end{equation}
A positive $H_N$ indicates that the efficiency-optimal operating point remains within the stable regime, while $H_N\leq 0$ indicates that optimization has become stability-constrained. We therefore define the \emph{Critical Model Size (CMS)} $N_{\mathrm{CMS}}$ as the model scale at which the optimal and safe learning rates coincide:
\begin{equation}
\eta_{\mathrm{opt}}(N_{\mathrm{CMS}})=\eta_{\mathrm{safe}}(N_{\mathrm{CMS}}).
\end{equation}
This provides a compact measure of how far a given architecture and training recipe can scale while simultaneously maintaining training stability and optimization efficiency, serving as a proxy metric for the model’s scaling stability.

\autoref{fig:max_safe_lr} shows the scaling trends of both learning rates observed in our experiments. For \modelname{}, $\eta_{\mathrm{safe}}$ remains substantially above $\eta_{\mathrm{opt}}$ across the measured scales. Extrapolating the fitted trends yields $N_{\mathrm{CMS}}^{\mathrm{act}}\approx 572$B and $N_{\mathrm{CMS}}^{\mathrm{total}}\approx 12$T, where $N_{\mathrm{CMS}}^{\mathrm{act}}$ and $N_{\mathrm{CMS}}^{\mathrm{total}}$ denote the CMS measured by activated and total parameters, respectively. These results indicate substantial stability headroom for scaling \modelname{} beyond the model sizes studied.

\subsection{Infrastructure}

\paragraph{Data Packing.} Sequence packing reduces padding overhead in language-model training~\citep{krell2021efficient}. To improve effective token throughput at a fixed sequence length and global batch size, we apply best-fit packing to the globally shuffled data stream using a bounded look-ahead window. The window contains up to $2BS$ tokens, where $B$ denotes the global batch size and $S$ the sequence length. Samples left unpacked in the current window are carried over and prioritized in the next iteration, preventing them from being indefinitely delayed by subsequent packing decisions. 
Long-context training amplifies compute imbalance arising from document-length variation. Packed sequences are therefore reordered across data-parallel replicas to balance computation at micro-batch granularity.

\paragraph{Sink Attention and Context Parallelism.} 
Context parallelism uses ring-based K/V exchange for full attention~\citep{liu2023ringattentionblockwisetransformers}, while sliding-window attention exchanges only the required halo and overlaps these exchanges with computation in both the forward and backward passes. During halo exchange, each rank computes attention over its local K/V, producing outputs for interior queries and partial outputs with associated Log-Sum-Exp (LSE) statistics for boundary queries. Once the halo arrives, the rank computes the complementary contributions for boundary queries and combines the local and halo partial outputs through LSE-based renormalization. This allows boundary queries to make progress before remote K/V becomes available while preserving sliding-window attention semantics. We illustrate the halo exchange schedule and partial output merging
in \autoref{fig:swa_halo_partial_output} in the appendix. Both full and sliding-window attention execution paths fuse learnable sink contributions into attention computation to reduce memory traffic.

\paragraph{Correctness Validation.} 
To reduce silent errors introduced by model and training changes during RSI, the agent mines semantic invariants from the model's training setup. These invariants are converted into checks that run during distributed training to detect deviations from single-machine semantics and violations of predefined correctness constraints.

%% file: sections/pre_training.tex
\section{Pre-Training}

\subsection{Pre-Training Data}
\label{sec:pretraining-data}
We construct a high-quality, multilingual pre-training corpus spanning diverse domains, drawing from web pages, public GitHub code, books, academic papers, news, and domain-specific materials. For the code, we preserve diversity across programming languages, modalities, and task granularities, dimensions emphasized by prior work on multilingual code generation, multi-granularity evaluation, and repository-level executable completion~\citep{ding2026nl2repobench,deng2025swebenchpro, khandpur2025swebenchmultilingual,chai2025multilingualmultimodalsoftwaredeveloper,xu2025m2gevalenhancingevaluatingmultigranularity,yang2024execrepobenchmultilevelexecutablecode}. Our unified data processing pipeline includes cleaning, quality filtering, exact deduplication, and fuzzy deduplication to reduce noise and redundancy while preserving diversity.

General pre-training proceeds in two stages with a progressively adjusted data mixture. 
\textbf{Stage 1} establishes foundational language capabilities and general knowledge through diverse corpora, maximizing unique token counts across multiple domains. \textbf{Stage 2} gradually increases the proportion of code and STEM data to strengthen programming, mathematical, and scientific problem-solving capabilities. Specifically, we design a four-phase pipeline increasing code data percentage from 40\% to 70\%.

Mid-training then combines high-quality natural corpora~\citep{starcoder2stackv2}, QA pairs~\citep{ahmad2025opencodeinstruct,opencodeinterpreter,NEURIPS2024_a4ca07aa,yang2025codesimpleqascalingfactualitycode}, and agentic trajectories~\citep{OctoPack} to prepare the model for downstream reasoning and agentic tasks. Across pre-training and mid-training, the model is trained on a total of 29T tokens.

\subsection{Hyper-Parameters}
\paragraph{Model Hyper-Parameters.}

\begin{table}[!t]
    \centering
    \small
    \caption{Summary of the main model architecture parameters.}
    \label{tab:pretraining-model-parameters}
    \setlength{\tabcolsep}{10pt}
    \renewcommand{\arraystretch}{1.22}
    \scalebox{0.9}{
    \begin{tabular}{lr}
        \toprule
        \textbf{Configuration} & \textbf{Value} \\
        \midrule
        Total Parameters & $320$B \\
        Activated Parameters & $15$B \\
        Transformer Layers & $88$ \\
        Hidden Dimension & $3{,}072$ \\
        Attention Heads (Q/KV) & $48/8$ \\
        Head Dimension & $128$ \\
        Hybrid Attention Pattern & $3$ SWA $+\ 1$ FA \\
        Sliding Window Size & $4{,}096$ \\
        Partial RoPE Dimensions & $32$ \\
        Experts (Total/Activated) & $256/8$ \\
        MTP Layers & \makecell{$2$ Independent (Training) \\ $1$ Recursive $\times 8$ (Inference)} \\
        MTP Sliding Window Size & $512$ (Inference) \\
        \bottomrule
    \end{tabular}}
\end{table}

As summarized in \autoref{tab:pretraining-model-parameters}, \modelname{} has $88$ Transformer layers and a hidden dimension of $3{,}072$. Attention uses $48$ query heads and $8$ key--value heads, each with a dimension of $128$. The first two and last two layers use full attention (FA) to capture global information and improve training stability, while the intermediate layers repeat three SWA layers followed by one FA layer, yielding a $3{:}1$ local-to-global ratio within each repeating block. The SWA uses a window size of $4{,}096$. Partial RoPE is applied to $32$ dimensions of each attention head, with RoPE bases of $10{,}000$ and $1{,}000{,}000$ for SWA and FA, respectively. Query-Aware Sink Attention is enabled in all attention layers.
The first Transformer layer uses a dense FFN with an intermediate dimension of $12{,}288$, while the remaining layers use sparse MoE FFNs. Each MoE layer contains $256$ routed experts, with $8$ experts activated per token and an intermediate dimension of $1{,}536$ for each expert. Both dense and expert FFNs adopt SwiGLU. The vocabulary size is $160{,}000$, with untied input embeddings and output weights.
During pre-training, \modelname{} adopts two Multi-Token Prediction (MTP) layers to improve the performance of the backbone model. Each prediction order uses an embedding dimension of $128$ and $8$ attention heads, and the MTP loss is weighted by $0.3$.

\paragraph{Training Hyper-Parameters.}

We employ Muon~\citep{liu2025muonscalable} for all parameters except token embeddings, the language modeling head, normalization parameters, and MoE routers, which are optimized with Adam. For Adam, we set $\beta_1 = 0.9$, $\beta_2 = 0.95$, and $\epsilon = 10^{-15}$. For Muon, we set the momentum to $0.95$. For an $m \times n$ weight matrix, we scale the orthogonalized update by $0.2\sqrt{\max(m,n)}$, combining spectral scaling with an additional factor of $0.2$ to align the update RMS with that of Adam-style updates~\citep{liu2025muonscalable}. We apply weight decay with a coefficient of $0.1$ to all parameters except layer normalization parameters, which are exempt from weight decay. We additionally apply QKNorm Clip, capping the affine parameters of QKNorm at $10$. We adopt a warmup-stable-decay learning rate schedule. The learning rate is warmed up to $6 \times 10^{-4}$ and held constant for the remainder of Stage 1, then decays to $2 \times 10^{-4}$ by the end of Stage 2 and to $6 \times 10^{-5}$ by the end of mid-training. 




\subsection{Mid-Training and Long-Context Extension}
\label{sec:midtraining}

\subsubsection{Training Curriculum}
\label{sec:midtraining-curriculum}

Mid-training bridges broad pre-training and specialized agent post-training through a curriculum that combines a diverse data mixture, selective token-level supervision, and progressive context extension.

\paragraph{Data Organization.}
The mid-training mixture comprises three principal categories: high-quality natural corpora, QA pairs, and agentic trajectories. We organize training data using a chat template adapted for mid-training, retaining a selected subset of special tokens, including those associated with reasoning and tool calls.


\paragraph{Token-Level Loss Masking.}
We specify loss masks directly for individual tokens, independently of the role-specific special tokens used to determine supervision in a typical post-training setup. This decouples data formatting from loss computation and allows supervision to be tailored to selected spans within each training example, with finer control than masking entire conversational roles.

\paragraph{Progressive Context Extension.}
We extend the maximum training context lengths through 32K, 128K, and 512K tokens, using variable-length sequence packing throughout. The global-attention RoPE base increases from $500,000$ at 32K to $1,000,000$ in the subsequent phases; the hybrid attention pattern, $4,096$ sliding window, and SWA RoPE base of $10,000$ remain fixed.

\paragraph{Optimization across Phases.}
The mid-training learning rate follows a single cosine decay across all three context-length phases, with no additional warmup. At phase transitions, we transfer model weights and reinitialize the optimizer state while preserving the learning-rate schedule's progress. 

\subsubsection{Agentic Data Construction Guided by Trajectories}
\label{sec:midtraining-synthetic-trajectories}
Agentic mid-training requires large and diverse trajectory corpora, but suitable trajectories are scarce and concentrated in a limited set of tasks and environments. To expand coverage while preserving realistic interaction patterns, we use selected trajectories as behavioral anchors and pre-training documents to broaden technical and task coverage.

Using these sources, we construct executable, containerized instances from compatible code repositories, following prior work on tasks derived from repositories and trajectory generation at scale~\citep{yang2025swesmithscalingdatasoftware,wu2026largescaleterminalagentictrajectory}. This process yields more than one million instances. We roll out an agent on each instance and retain valid and complete trajectories for mid-training.

\subsection{Evaluation}
\label{sec:midtraining-eval}

We evaluate pre-training and mid-training using two complementary types of evaluations. First, we compute \textit{bits-per-byte (BPB) on held-out agent trajectories}. Lower BPB indicates that the model assigns higher likelihood to behaviors observed in competent agent trajectories, providing a generation-free measure of how well the model captures the trajectory distribution. This metric can provide a fine-grained training signal that complements downstream task-level evaluations.

We further decompose long-horizon agent behavior into \textit{atomic abilities} and evaluate how reliably the model performs each ability. This fine-grained evaluation helps identify specific capability gaps and informs data selection and curriculum design. Together, these two evaluation paradigms provide complementary signals for tracking and improving agentic capabilities during pre-training and mid-training.



%% file: sections/post_training.tex
\section{Post-Training}


\subsection{Large-Scale and Diverse Task Synthesis}
\label{sec:task-env-synthesis}

\begin{figure}[t]
    \centering
    \includegraphics[width=\linewidth]{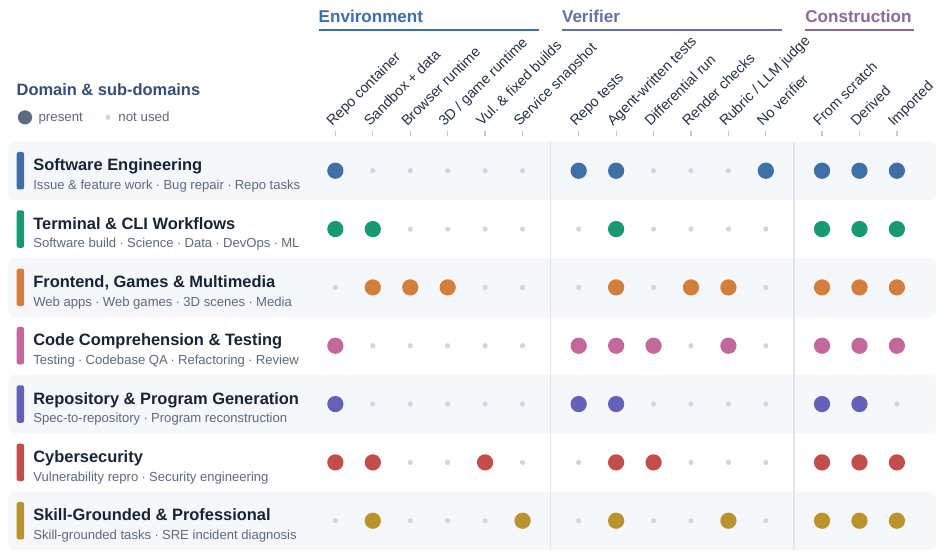}
    \caption{Overview of coding-agent task instances and their environments, verifiers, and construction methods.}
    \label{fig:coding-agent-instances}
\end{figure}

We organize the task synthesis framework into three settings: general, reasoning, and coding. 
Each setting uses diverse data sources and a domain-specific pipeline to generate tasks at scale.

\paragraph{General Agent.}
\label{sec:general-agent}
We collect APIs and MCP servers from RapidAPI\footnote{\href{https://rapidapi.com}{https://rapidapi.com}} and Smithery\footnote{\href{https://smithery.ai}{https://smithery.ai}}, public skills from the web, and workspace files for task synthesis. Guided by the Standard Occupational Classification \cite{us2018standard}, we cover a broad range of occupations and synthesize high-coverage environments and tasks, which are iteratively refined through a solve-then-refine loop.

\paragraph{Reasoning Agent.}
\label{sec:reasoning-agent}
We synthesize tasks through a document-grounded pipeline that jointly controls concept coverage, material quality, and task difficulty. The pipeline first expands a broad concept inventory into fine-grained scientific topics, then retrieves and curates documents for each topic, filters the documents before synthesis, and finally converts the retained materials into tasks with reference answers. By covering high-value knowledge through the document collection, the pipeline promotes corresponding coverage in the synthesized prompts.

\paragraph{Coding Agent.}
\label{sec:coding-agent}
We construct executable environments from GitHub repositories and industrial data~\cite{jimenez2024swe,pan2025training,yang2025swesmithscalingdatasoftware,deepseekai2026deepseekv4,yang2026incoder32bcodefoundationmodel,loopcoder,loopcoderv2} using multiple versions of coding-agent harnesses such as Claude Code and Codex, with our model driving environment construction and skills supporting setup. Task queries draw on PRs and issues, repository knowledge graphs and documentation, and reusable patterns extracted from cleaned, de-identified usage logs. Skills capture task structure and common failure modes to guide instance synthesis. Verifiers combine agent-written tests, adapted repository tests, and checks derived from documented behavior. For offline instances, external dependencies are replaced by local fixtures or mocks. An agent-judge loop checks executability, leakage, and consistency among instructions, reference solutions, environments, and verifiers, repairing defective instances and rejecting weak assertions. Reference solutions must pass while no-op solutions must fail. Rubric-based trajectory review further filters cheating and low-quality execution beyond binary success. Repeated agent runs estimate difficulty, and overly easy instances are enriched with additional skills or constraints before revalidation; valid instances exposing model failures are retained.

An overview of coding-agent task instances is shown in \autoref{fig:coding-agent-instances}. We construct instances through two complementary modes: building new environments from raw sources and updating verified environments through re-tasking, rewriting, fault injection, evolution, hardening and repair, and selective re-rolling. The resulting library covers software engineering, interactive and multimodal tasks, cybersecurity, and skill-grounded professional work, with domain-specific runtimes and verifiers.

\begin{figure*}[t!]
    \centering
    \includegraphics[width=1\textwidth]{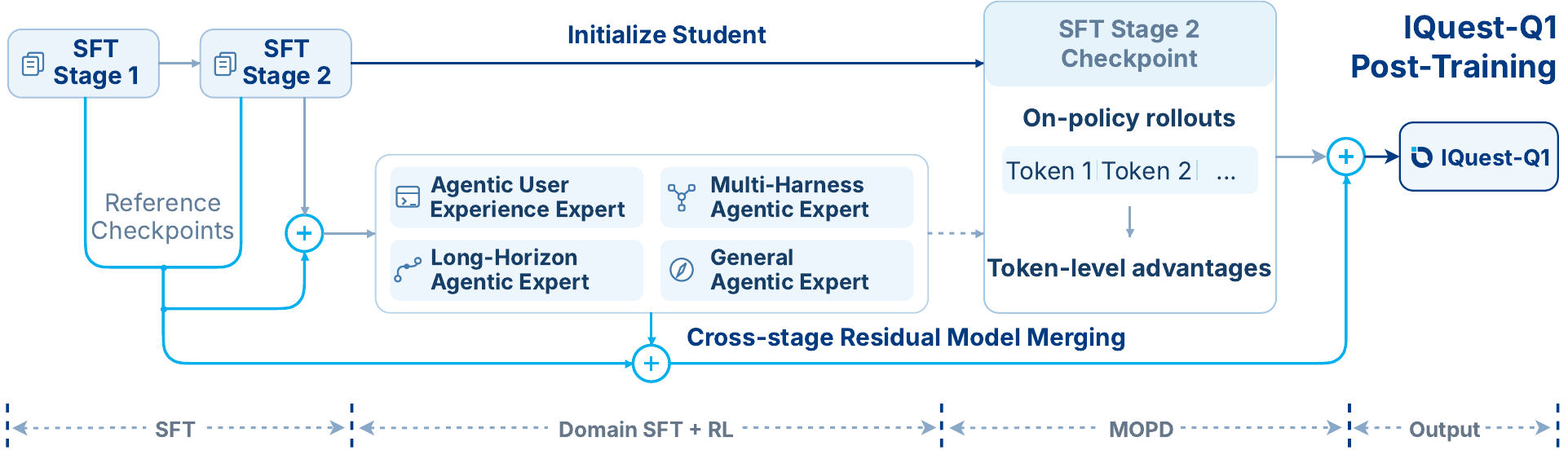}
    \caption{\modelname{} Post-Training Pipeline.}
    \label{fig:IQuest-Q1-Post-Training}
\end{figure*}

\subsection{Environment Integrity and Reward Hacking Mitigation}
Agentic training environments are vulnerable to reward hacking, where models exploit leaked solutions or weaknesses in task verifiers instead of solving the intended tasks. Such behavior can contaminate SFT data and introduce misleading reward signals during RL. To prevent this, we apply three layers of protection: environment preparation, execution-time checks, and trajectory auditing.

\paragraph{Environment Preparation.}
Before collecting trajectories, we prepare task environments to prevent solution leakage. Depending on the task, we clean Git history, remove files that could reveal existing solutions, or create fresh environments containing only the required materials. We also isolate task environments to prevent unintended access to hidden evaluation data or resources from other tasks.

\paragraph{Execution-Time Checks.}
During rollouts, we restrict access to external resources and sensitive task files. Depending on the environment, we enforce network restrictions or intercept tool calls before execution. We develop blocking rules from observed hacking behaviors to prevent agents from retrieving existing solutions, upstream code, or hidden evaluation materials. These checks preserve access to legitimate dependencies and task resources. Blocked operations return explicit feedback, allowing agents to continue without the prohibited access.

\paragraph{Trajectory Auditing.}
We further audit collected trajectories using rule-based checks and rubric-guided LLM judgments. The audits examine executed commands, tool outputs, and submitted solutions to identify behaviors such as retrieving leaked fixes, accessing hidden tests, or modifying evaluation scripts. Confirmed hacking trajectories are excluded from training, while suspicious cases can be flagged for further review. We also use detected cases to refine our environment restrictions and blocking rules.

\paragraph{Task-Specific Safeguards.}
We adapt these safeguards to the needs of each task. For SWE-like tasks, we keep only the Git history up to the base commit and use hooks in Claude Code, Codex, and OpenCode to block attempts to retrieve existing solutions. For Terminal-Bench, we remove Git history when it is not needed and check commands before execution when it must be preserved. ProgramBench takes a different approach, building fresh containers with only the compiled oracle and task documentation, without exposing the original repository.

\subsection{Supervised Fine-Tuning}
\label{sec:sft}

As shown in \autoref{fig:IQuest-Q1-Post-Training}, we initialize supervised fine-tuning (SFT) from the mid-trained checkpoint to prepare the policy for reinforcement learning. We optimize the autoregressive cross-entropy loss on assistant responses, including tool-call tokens, while masking user instructions and tool observations from the loss. SFT proceeds in two stages (240B+ tokens in \textbf{Stage 1} and 50B+ tokens in \textbf{Stage 2}), with initial learning rates of $5\times10^{-5}$ and $1.5\times10^{-5}$, respectively, and cosine learning-rate decay within each stage. Multiple samples are packed into each sequence, with attention masking to prevent cross-sample attention.

\input{sections/reinforcement_learning}

\subsection{Multi-Teacher On-Policy Distillation}
\label{sec:mopd}

As shown in \autoref{fig:IQuest-Q1-Post-Training}, we initialize the student from SFT Stage 2 and distill knowledge from four
frozen expert models:
(1) an \emph{agentic user experience expert} for terminal
interaction, tool use, web development, and coding workflows;
(2) a \emph{multi-harness agentic expert} for operation across
Claude Code, Codex, and OpenCode;
(3) a \emph{long-horizon agentic expert} for extended coding
and cybersecurity tasks; and
(4) a \emph{general agentic expert} for working-agent tasks,
web search, multi-turn conversation, and reasoning.

Following MOPD~\citep{ma2026mopdmultiteacheronpolicydistillation,
kimiteam2026kimik3openfrontier}, we collect $B$ rollouts from
$p_k=p_{\theta_k}$ and select a frozen teacher $q_{d_i}$ for
each task $x_i$ according to its domain, $d_i=d(x_i)$.
Given the interaction context $h_{i,t}$, the teacher evaluates
student-generated tokens to obtain a clipped distillation advantage:
\begin{equation}
\tilde a_{i,t}
=
\operatorname{clip}\!\left(
\log\frac{q_{d_i}(y_{i,t}\mid h_{i,t})}
         {p_k(y_{i,t}\mid h_{i,t})},
-c,c\right),
\quad c>0.
\label{eq:mopd-adv}
\end{equation}
Motivated by reverse-KL distillation~\citep{gu2024minillm},
we optimize the length-normalized surrogate:
\begin{equation}
L_k(\theta)
=
-\frac{1}{B}\sum_{i=1}^{B}\frac{m_i}{T_i}
\sum_{t\in\mathcal A_i}
\tilde a_{i,t}\log p_\theta(y_{i,t}\mid h_{i,t}),
\label{eq:mopd-loss}
\end{equation}
where $\mathcal A_i$ contains model-generated token positions,
including tool-call tokens, and $T_i=|\mathcal A_i|$.
Instructions and tool observations provide context but are
excluded from the loss.
We set $m_i=0$ for repetitive rollouts and $m_i=1$ otherwise.
Rollouts, contexts, advantages, and masks are fixed during
differentiation.

\subsection{Model Merging}
As shown in \autoref{fig:IQuest-Q1-Post-Training}, starting from SFT Stage 1, we interleave training with residual model merging across stages and within expert branches, ultimately incorporating expert and MOPD checkpoints into \modelname{}. Each merge combines parameter deltas relative to a designated reference checkpoint using prescribed weights and global $\ell_2$-norm calibration, then adds the resulting update back to the reference parameters.

\subsection{Post-Training Infrastructure}
\label{sec:post-training-infra}

\paragraph{Multi-harness and Multi-version Agent Runtime.}
A single policy can be trained on trajectories from multiple harnesses and versions within one run.
Each training instance specifies its scaffold, task environment, and verifier. The runtime executes the corresponding configuration and verifies the task in an isolated sandbox while preserving tool orchestration, context management, and interaction flow. We map different API protocols to a unified policy interface to reduce the difficulty of downstream integrations.

\paragraph{Rollout-training Consistency.}
We implement an end-to-end TITO path at the model-serving boundary. Specifically, we apply a session-level prefix tree to retain token histories shared across turns and branches. 
A request then combines its environment observations with a recorded token prefix. The corresponding response is returned in the structured format expected by the scaffold. Finally, we retain trajectories paired with rewards and token-level masks from the tree at the end of the sequence.


\paragraph{Asynchronous Distillation and Teacher Execution.}
We implement MOPD in a fully asynchronous training system. We apply periodic synchronization of student parameters and explicit control of sample staleness, limiting the policy lag introduced by asynchronous execution. We execute teacher forward passes with the training engine on trajectories generated online by the student, producing token-level teacher log probabilities. 
We use the same execution path for teacher evaluation and student log-probability computation to reduce discrepancies caused by the training engine.

This framework supports independently configured teachers through a common scheduling interface. Teachers reuse the same execution resources through weight switching. We store teacher weights in distributed storage and load them on demand, with only the shards required by the active teacher placed in GPU memory. 
Weights of the next teacher are prefetched during the current forward pass with overlapping computation. This design supports additional teachers without a proportional increase in resident GPU replicas.

%% file: sections/reinforcement_learning.tex
\subsection{Environment-Centered Reinforcement Learning}
\label{sec:environment-rl}

We outline an RL design that targets verifiable task completion while accounting for user constraints and execution cost. It draws on configurable-harness training in Kimi K3~\citep{kimiteam2026kimik3openfrontier} and domain-specialized Group Relative Policy Optimization (GRPO) in DeepSeek-V4~\citep{deepseekai2026deepseekv4}.

\paragraph{Multi-harness, Multi-version Episodes.}
To improve the model's performance across different agent harnesses and versions, we train a shared policy on trajectories collected from diverse harness configurations, including Claude Code, Codex, and their different versions. 
Each episode specifies a task domain, harness and its version, environment, and evaluator.
During training, the policy interacts with different harnesses, observing public instructions, tools, and execution feedback while reference solutions and hidden checks remain isolated.
GRPO compares repeated rollouts of the same task under a fixed harness and evaluator version. Sampling balances domain--harness combinations and retains regression cases.


\paragraph{Rewards and Failure Attribution.}
Independent verifiers determine terminal correctness. For long trajectories, auxiliary segment-level rewards are restricted to measurable progress, while penalties address unnecessary computation, constraint violations, and reward hacking. Model-generated diagnoses require evidence from execution records before contributing to these signals. Failures are attributed before optimization: policy-caused errors supply learning signals, environment or harness faults unrelated to policy behavior are masked, and evaluator faults block sample consumption. Token-level updates use recorded rollout probabilities through the existing Token-In, Token-Out (TITO) pipeline. The resulting specialized policies can subsequently be consolidated through MOPD.

%% file: sections/evaluation.tex
\section{Evaluation}\label{sec:evaluation}
\subsection{Benchmarks}
We evaluate \modelname{} on a comprehensive suite of benchmarks spanning two broad capability axes, covering both coding and general agentic capabilities:
\begin{itemize}[leftmargin=*]
    \vspace{-4mm}
    \item \textbf{Coding Agent}: DeepSWE v1.1~\cite{huang2026deepswebenchmark,datacurve2026deepswev11}, SWE-bench Multilingual~\cite{khandpur2025swebenchmultilingual,yang2025swesmithscalingdatasoftware}, SWE-Bench Pro~\cite{deng2025swebenchpro}, NL2Repo-Bench~\cite{ding2026nl2repobench}, ProgramBench~\cite{yang2026programbench}, CyberGym~\cite{wang2026cybergym}, SkillsBench~\cite{li2026skillsbench}, Terminal-Bench 2.1~\cite{merrill2026terminalbench,harbor2026terminalbench21},
    Terminal-Bench 3.0~\cite{merrill2026terminalbench,terminalbench2026v30}, Terminal-Bench 4.0~\cite{merrill2026terminalbench, marten2026terminalbench4},     
PaperBench~\cite{starace2025paperbenchevaluatingaisability}, LiveCodeBench v6~\cite{jain2024livecodebench,livecodebench2025v6}, IQuest-CLIBench.
    \item \textbf{General Capabilities}: 
    GPQA Diamond~\cite{rein2024gpqa}, 
    Humanity's Last Exam~\cite{phan2025humanitysexam},
    GDPval~\cite{patwardhan2025gdpval}, APEX-Agents v1.1~\cite{bennett2026apexagents11}, Toolathlon-Verified~\cite{li2025toolathlon}, MCP-Atlas~\cite{bandi2026mcp}, Claw-Eval~\cite{ye2026claweval}, AutomationBench~\cite{shepard2026automationbench}, JobBench~\cite{li2026jobbench}, Agents' Last Exam (ALE)~\cite{sun2026agentslastexam}, $\tau^3$-Banking~\cite{shi2026tauknowledge,sierra2026tau3bench}, BrowseComp~\cite{wei2025browsecomp}, WideSearch~\cite{wong2025widesearch},
    Humanity's Last Exam~(w/ tools)~\cite{phan2025humanitysexam},
    HealthBench Professional~\cite{hicks2026healthbenchprofessional}, IMOAnswerBench-Verified~\cite{luong2025imobench}, 
    CritPt~\cite{zhu2025probing}.
\end{itemize}

IQuest-CLIBench is an in-house benchmark for evaluating CLI agents in human-in-the-loop settings through script-driven user simulation, using both execution traces and environment-level state invariants. We integrate it into the core evaluation suite of \modelname{} to assess user experience and interaction quality alongside conventional task performance.

\begin{table}[!t]
    \centering
    \small
    \setlength{\tabcolsep}{2.8pt}
    \renewcommand{\arraystretch}{1.05}
\caption{Performance across coding and general capability benchmarks.
Scores are reported at the highest available reasoning effort for each model.
DS denotes DeepSeek. 
ProgramBench reports raw pass rates. PaperBench is evaluated in the CodeDev setting.
IQuest-CLIBench$^\ddagger$ is our proprietary built-in benchmark for evaluating human-agent collaboration and user experience (see Section \ref{sec:evaluation} for details). 
Results with * are from our own testing.
}
    \resizebox{\textwidth}{!}{
    \begin{tabular}{@{}l*{7}{>{\centering\arraybackslash}p{1.8cm}}@{}}
    \toprule
        \textbf{Benchmark} &
        \makecell{\textbf{\modelname{}}} &
        \makecell{\textbf{GLM-5.3}\\\textbf{Flash}} &
        \makecell{\textbf{DS-V4}\\\textbf{Flash-0731}} &
        \makecell{\textbf{Tencent}\\\textbf{Hy3}} &
        \makecell{\textbf{MiniMax}\\\textbf{M3}} &
        \makecell{\textbf{Claude}\\\textbf{Opus-5}} &
        \makecell{\textbf{GPT-5.6}\\\textbf{Sol}} \\
    \midrule
        \rowcolor{iquestazure!12}
        \multicolumn{8}{@{}l@{}}{\textbf{Coding Capabilities}} \\
        DeepSWE v1.1 & 64.6 & 63.4 & 54.4 & 28.0 & - & 73.7 & 73.0\\
        SWE-bench Multilingual & 80.3 & 79.7* & 78.7* & 75.8 & 75.0* & 89.5 & 74.1\\
        SWE-Bench Pro & 57.5 & 59.4* & 57.3* & 57.9 & 59.0 & 79.2 & 64.6  \\
        NL2Repo-Bench & 63.0 & 56.3 & 54.2 & 45.6 & 42.1 & 75.3 & 56.8 \\
        ProgramBench & 60.9 & 58.5* & 62.1* & 47.4* & 33.8* & 82.3 & 77.6 \\
        CyberGym & 84.5 & 81.6* & 76.7 & 51.8 & 40.3* & - & 83.6 \\
        SkillsBench & 74.1 & 56.2* & 72.0* & 63.9* & 54.2* & 76.2* & 71.5* \\
        Terminal-Bench 2.1  & 83.2 & 84.3 & 82.7 & 71.7 & 66.0 & 89.1 & 88.8\\
        Terminal-Bench 3.0 & 17.9 & 30.0* & 7.6 & 3.0* & 1.7* & 42.7 & 34.6\\
        Terminal-Bench 4.0 & 22.7 & 32.8 & 7.0 & 4.5* & 2.0 & 53.9 & 37.3 \\
        IQuest-CLIBench$^\ddagger$ & 53.7 & 49.5* & 51.5* & 36.6* & 30.7* & 58.2* & 58.5*\\
        PaperBench & 86.0 & 86.5* & 85.8* & 68.4* & 67.7* & 91.8* & 90.5\\
        LiveCodeBench v6 & 89.7 & 88.0* & 90.6 & 91.0 & 81.1* & - & - \\
    \midrule
        \rowcolor{iquestazure!12}
        \multicolumn{8}{@{}l@{}}{\textbf{General Capabilities}} \\
        GPQA Diamond & 93.3 & 92.2 & 89.9 & 90.9 & 92.9 & 93.4 & 94.6\\
        Humanity's Last Exam & 39.2 & 39.9 & 38.6 & 33.5 & 39.0 & 54.9 & 49.5 \\
        Humanity's Last Exam~(w/ tools) & 55.7 & 55.3 & 51.5 & 53.2 & 41.2* & 60.9 & 60.2 \\
        CritPt & 18.3 & 15.4 & 16.6 & 4.9 & 3.7 & 29.1 & 32.3\\
        IMOAnswerBench-Verified & 93.0 & 75.3* & 92.8* & 92.8* & 79.6* & - & -\\
        GDPval v2.1 & 1530 & 1641 & 1427 & 1045 & 1230 & 1708 & 1588 \\
        Toolathlon-Verified & 75.0 & 78.4 & 70.3 & 56.2 & 53.7* & 80.6 & 74.9 \\
        MCP-Atlas & 78.4 & 75.6 & 72.8* & 79.1 & 74.2 & 85.8 & 81.8\\
        Claw-Eval & 77.1 & 74.2* & 68.8* & 63.5* & 69.8* & 79.6* & 70.2* \\
        AutomationBench  & 52.7 & 48.8 & 37.7 & 16.1 & 19.7* & 50.3 & 45.8\\
        APEX-Agents 1.1 & 50.9 & 52.8 & 36.9* & - & - & 65.8 & 51.4 \\
        JobBench & 55.7 & 49.7* & 50.0* & 34.6 & 40.1* & 65.7 & 45.4\\
        Agents' Last Exam & 29.6 & 26.3 & 25.2 & 17.1 & 9.1* & 32.2 & 30.6 \\
        $\tau^3$-Banking & 51.8 & 47.2 & 39.4 & 22.9 & 15.3 & 44.7 & 44.3 \\
        BrowseComp & 85.5 & - & 78.0* & 84.2 & 83.5 & 90.8 & 90.4\\
        WideSearch & 81.8 & - & 80.5 & 81.9 & 73.5* & 84.0 & 86.3 \\
        
        HealthBench Professional & 33.4 & 47.1* & 26.4* & 29.8* & 24.7* & 59.8 & 60.5 \\

    \bottomrule
    \end{tabular}
    }
    \label{tab:agentic_combined}
\end{table}

\subsection{Baselines}

We compare \modelname{} with proprietary and open-source models. For proprietary models, we compare against Claude Opus 5~\cite{anthropic2026claudeopus5} and GPT-5.6 Sol~\cite{openai2026gpt56systemcard}. For open-source models, we compare with GLM-5.3-Flash~\cite{zai2026glm53flash}, DeepSeek-V4-Flash-0731~\cite{deepseekai2026deepseekv4}, Hy3~\cite{tencent2026hy3}, and MiniMax M3~\cite{minimax2026m3}.

\subsection{Evaluation Configurations}
For each model, we report a publicly available score when one is available; otherwise, we evaluate the model using the corresponding benchmark setup:
(1) \emph{Harness:} For agentic coding tasks, we use mini-SWE-agent~\cite{sweagent2025minisweagent} for DeepSWE v1.1, BasicAgent~\cite{starace2025paperbenchevaluatingaisability} for PaperBench, and Claude Code~\cite{anthropic2025claudecode} for other tasks. For general agentic tasks, we use our internal harness for search agent tasks and Humanity's Last Exam, and the benchmark-provided harnesses for other tasks.
(2) \emph{Runtime:} We set six-hour limits for CyberGym~\cite{wang2026cybergym} and ProgramBench, eight hours for Terminal-Bench 2.1~\cite{harbor2026terminalbench21}, ten hours for Terminal-Bench 3.0~\cite{terminalbench2026v30} and 4.0~\cite{marten2026terminalbench4}, and 12 hours for PaperBench~\cite{starace2025paperbenchevaluatingaisability}.

\subsection{Results}
\label{sec:evaluation_results}

\autoref{tab:agentic_combined} shows the results under the reported evaluation configurations. We highlight representative results and remaining gaps below.

\paragraph{Coding and Software Engineering.}
\modelname{} demonstrates a broad range of repository-level coding capabilities, including bug fixing and feature addition, achieving scores of 57.5 on SWE-Bench Pro and 64.6 on DeepSWE v1.1.
It also handles repository generation from natural language across NL2Repo-Bench (63.0) and PaperBench (86.0), covering general programming and research code reproduction.
Terminal-Bench and SkillsBench further evaluate long-horizon terminal workflows and skill-driven coding. On our internal IQuest-CLIBench, \modelname{} further demonstrates practical CLI interaction capabilities.



\paragraph{General Capabilities.}
\modelname{} scores 75.0 on Toolathlon-Verified, 52.7 on AutomationBench, and 77.1 on Claw-Eval, reflecting its capabilities in tool use and multi-step task execution. On BrowseComp and WideSearch, it reaches 85.5 and 81.8, respectively, demonstrating information-seeking capabilities across the two search settings. It also records 55.7 on JobBench and 50.9 on APEX-Agents 1.1, covering task execution in professional workflows. Beyond agentic interaction, scores of 93.3 on GPQA Diamond and 93.0 on IMOAnswerBench-Verified reflect its scientific and mathematical reasoning capabilities. Expert-level reasoning remains challenging: HLE scores of 39.2 without tools and 55.7 with tools indicate room for improvement in both settings.

%% file: sections/case_studies.tex
\section{Capabilities with CLI Agents}
\label{sec:case_studies}

\begin{figure}[!t]
    \centering
    \includegraphics[width=0.9\linewidth]{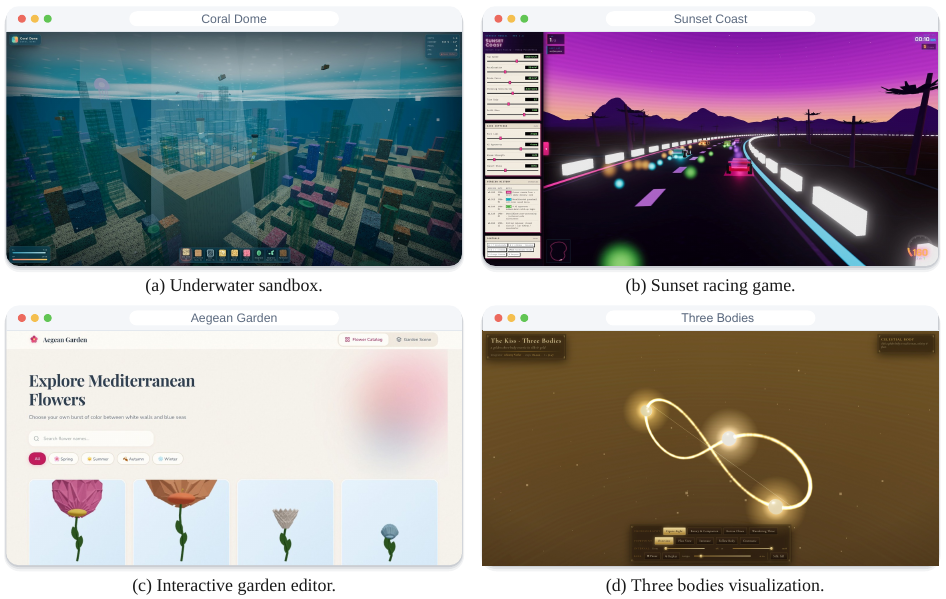}
    \caption{Representative interactive frontend applications generated by \modelname{} from natural-language specifications, covering game worlds, scene editing, and scientific exploration.}
    \label{fig:interactive_frontend}
\end{figure}

\subsection{Interactive Frontend Generation}
\label{sec:interactive_frontend}

\modelname{} generates diverse interactive frontend applications from user queries, combining visual design, scene construction, and interaction logic in browser-runnable HTML, CSS, and JavaScript. These tasks showcase its CLI coding capabilities by turning implementations into interactive artifacts.

We provide examples (see \autoref{fig:interactive_frontend}) from domains including games, creative tools, and scientific visualization. For example, the underwater sandbox combines an explorable world with building and mining, while the sunset racing game couples vehicle control with camera movement and race-state feedback. The garden editor links flower parameters and planting actions to a 3D scene. Together, they suggest how \modelname{} coordinates rendering, user input, and application state to create interactive web experiences with distinct visual styles.

\subsection{Iterative Development under Human Supervision}

We illustrate a model development iteration in which an agent built on \modelname{} identifies a capability gap, implements a solution, and evaluates the model under human supervision and intervention.
The iteration targets adaptation to different scaffolds and tool-call schemas. While the model handles familiar schemas well, it can fail on unfamiliar ones, producing 35 tool-call errors in one long-horizon task due to mismatched tool names and parameters.

\textbf{Trace Tool-call Failures and Identify Data Inconsistencies.}
The agent traces the failures across 6,054 tool calls and finds that the training data is heavily biased toward a single tool-call schema, with some records containing mismatches between tool definitions and call parameters due to preprocessing. It proposes cleaning these inconsistent records and augmenting the training data with diverse tool-call schemas.
Human experts review the diagnosis and proposed intervention before approving the training run.

\textbf{Implement Mixed-schema Data Augmentation and Model Training.}
The agent implements the data cleaning and augmentation pipeline and prepares the experiment configuration. The training run uses 2.35B tokens and 938.7 GPU hours. The resulting model checkpoint is evaluated on held-out schemas and familiar schemas, while checking for performance regressions.

This iteration demonstrates the potential of iterative model development with \modelname{}, while also highlighting the importance of human supervision.
The agent drives failure analysis, intervention design, implementation, and evaluation, with human experts reviewing key decisions and controlling costly experiments and changes carried forward to subsequent iterations.

\subsection{Cyber Security}
Beyond general software engineering tasks, we further evaluate \modelname{} on defensive cybersecurity tasks, where understanding and validating vulnerabilities are essential for improving software security. \modelname{} achieves a score of 84.5 on CyberGym~\cite{wang2026cybergym}, a benchmark that evaluates the ability to reproduce and validate real-world vulnerabilities in large open-source codebases, providing evidence to support vulnerability diagnosis and remediation.

We further assess the practical utility of these capabilities in real-world security workflows. During a three-day deployment, human cyber security experts, assisted by \modelname{}, produced 30 vulnerability reports. All reports were independently validated by fellow security experts, and findings affecting open-source projects were responsibly disclosed to their respective maintainers.
The participating experts reported that \modelname{} provided substantial support for their defensive security work, particularly in understanding and validating security issues. 


%% file: sections/limitations.tex
\section*{Limitations}


\begin{itemize}[leftmargin=*]
    \item Text-only input. \modelname{} has no native image, audio, or video input capability.
    \item Output reliability. Generated explanations and code may be incorrect. Code changes should be reviewed or verified with task-appropriate tests.
    \item Tool integration. Tool calls use the IQuest-specific format in the chat template. Structured tool execution and reasoning extraction require compatible serving parsers and an agent harness.
    \item Challenges in Real-World CLI Tasks. Real-world CLI tasks often require iterative debugging and verification. \modelname{} may overlook constraints, repeat failed attempts, or leave issues unresolved, necessitating human review and supervision.
    \item Ongoing development. \modelname{} still has limitations in its capabilities and reliability, and we will continue to advance its capabilities. Furthermore, we release the model weights, deployment configurations, and agent-integration instructions to facilitate further research and continued development in the community.
\end{itemize}




%% file: sections/conclusion.tex
\section*{Conclusion}
We present \modelname{}, an open-source agentic foundation model with broad CLI capabilities. We train \modelname{} through a multi-stage pipeline to progressively develop general knowledge, long-context capabilities, and agentic CLI capabilities. \modelname{} demonstrates capabilities on coding, general agentic tasks, and reasoning-heavy tasks. 
Furthermore, \modelname{} participates in our internal iterative model-development loop, 
providing a foundation for continuing model development.




%% file: sections/author_list.tex
\section{Contributions and Acknowledgements}
\textbf{The authors of this paper are listed follows:} 
\begingroup
\renewcommand{\thefootnote}{}
\renewcommand{\footnoterule}{}
\footnotetext{\noindent
$^{*}$ Core contributors; $^{\dagger}$ Senior advisory committee; 
$^{\ddagger}$ Corresponding author.}
\endgroup
A. Jian Yang$^{*}$, An Huang$^{*}$, Bo Dan$^{*}$, Chen Mu$^{*}$, Chengran Yang$^{*}$, Chuan Shi$^{*}$, Feng Chang$^{*}$, Hongxin Li$^{*}$, Hongyuan Tao$^{*}$, Jinxing Zhang$^{*}$, Jujie He$^{*}$, Kun Wang$^{*}$, Lin Jing$^{*}$, Ming Yang$^{*}$, Qicheng Hu$^{*}$, Renyuan Li$^{*}$, Shark Liu$^{*}$, Shawn Guo$^{*}$, Shi Qing$^{*}$, Shi Zhong$^{*}$, Tianyu Zheng$^{*}$, Wei Liu$^{*}$, Wei Zhang$^{*}$, Wenrui Cai$^{*}$, Wenxiao Chen$^{*}$, Xuxing Chen$^{*}$, Yan Xing$^{*}$, Yi Xing$^{*}$, Yifan Luo$^{*}$, Yiran Xing$^{*}$, Yizhi Li$^{*}$, Yongsheng Kang$^{*}$, Yucheng Ding$^{*}$, Yue Song$^{*}$, Yun Qing$^{*}$, Yunqi Liu$^{*}$, Zelong Huang$^{*}$, Zhengmao Ye$^{*}$, Zhentao Fan$^{*}$, Zhiteng Li$^{*}$, Ziling Huang$^{*}$,
Bihong Tang, Bingli Wang, Ce Jun, Changhao Chen, Chen Hu, Chen Xiao, Chi Liu, Chunjiang Ge, Danyang Li, Deyang Kong, Dian Qi, Ding Tan, Evan Yang, Fanglin Xu, Fei Bai, Gang Liu, Guochang Li, Hao Tang, Haoran Feng, Haowen Wang, Huahao Hu, Huanjin Yao, Huatong Song, Jia Deng, Jiajun Wu, Jialiang Cheng, Jiayi Liang, Jiayi Wu, Jie Deng, Jiejun Tan, Jing Yang, Jingda Xu, Junhang Cheng, Junxiao Liu, Kai Lv, Kaixiang Pan, Kecheng Lu, Kun Zeng, Kun Zheng, Kunbo Ding, Liang Qiu, Lingyi Sun, Luan Zhang, Mengxiao Lv, Ning Dong, Pei Wang, Peng Zhang, Qingsong Cai, Qixiang Yin, Qiyao Wang, Renbiao Liu, Renfei Dang, Ruihang Li, Ruirui Huang, Shanhui Yin, Shen Jian, Shengjie Fang, Shikai Li, Shinan Zhang, Shiqi Chen, Shuai Liu, Shuangqi Li, Shuo Tang, Shuqing Deng, Sijin Hao, Siwei Wu, Siyue Wang, Tao Hai, Tao Ren, TingYang Chen, Wei Shao, Weicheng Gu, Weihan Peng, Wenlong Miao, Xavier Yu, Xiangyuan Ji, Xiaocheng Li, Xiaodong Huang, Xiaohan Ge, Xiaoxiang Shi, Xiaoyang Zhang, Xiaoyu Yang, Xidong Wang, Xin Li, Xinyu Tang, Xiyuan Yang, Xudong Ma, Ya Tong, Yachao Wang, Yang Tian, Yaxin Du, Yeli Ruan, Yi Wang, Yi Yu, Yibin Feng, Yibo Tan, Yihang Wang, Yiming Ju, Yinda Chen, Yinda Chen, Yingjie Du, Yiwei Liu, Yizhi Zhou, Yuan Wei, Yuli Chen, Yuqi Zhu, Yuwen Li, Yuxi Dong, Yuxuan Zhang, Yuyang Song, Zhange Zhang, Zhaohan Zhang, Zhaoyi Li, Zhen Zhang, Zhengbo Zhang, Zhihang Hu, Zhiwei Yu, Zhixin Xie, Zhiyu Chen, Zhizhao Luo, Zhongyuan Wu, Zhou Fang, Zhu Lin, Zhuocheng Gong, Zixi Zhao, Ziyang Zeng, Ziyi Wang, Ziyu Zhao,
Ran Tao$^{\dagger}$, Mingjie Tang$^{\dagger}$, Chuan Hao$^{\dagger}$, Peihao Wu$^{\dagger}$, Jianzhou Wang$^{\dagger}$, Nan Jiang$^{\dagger}$, Yuan Zhi$^{\dagger}$, Bryan Dai$^{\ddagger}$.

\clearpage

%% file: sections/appendix.tex
\newpage

\appendix

\section{Pre-Training}
\label{app:pre-training}

\begin{figure}[t]
  \centering
  \includegraphics[width=0.7\linewidth]{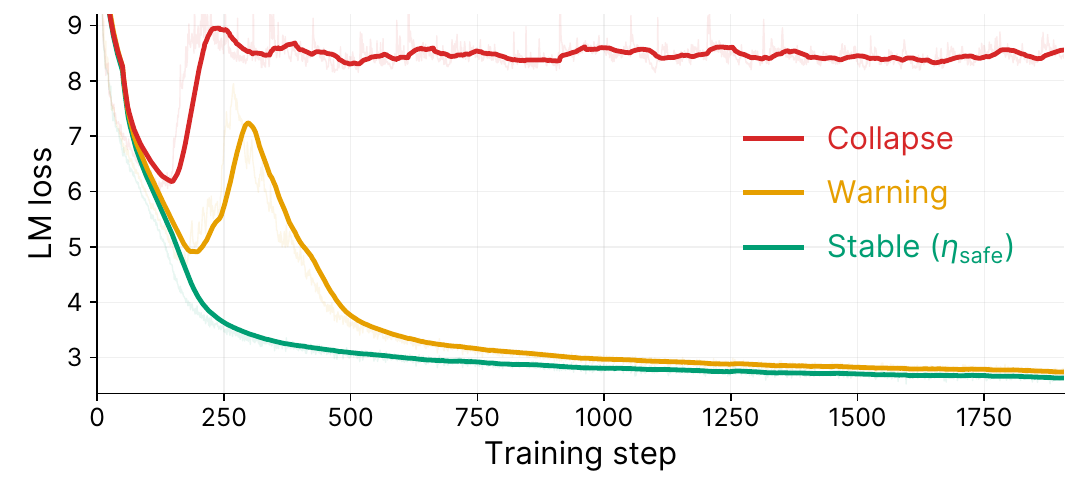}
  \caption{
    Representative training-loss trajectories for the model under stable, warning, and collapsed training regimes. 
  }
  \label{fig:safelr-definition}
\end{figure}

\paragraph{Safe learning rate.}\label{app:pre-training:safe-lr} We perform a grid search over learning rates and classify the resulting training-loss trajectories as stable, warning, or collapsed, as illustrated in \autoref{fig:safelr-definition}. Stable runs show consistently decreasing loss; warning runs exhibit a transient rebound before recovering, while collapsed runs rebound and remain at high loss. We define the safe learning rate, $\eta_{\mathrm{safe}}$, as the largest tested learning rate classified as stable.

\paragraph{$\mu$P scaling factors.}
\autoref{tab:mup_scaling_factors} summarizes the width- and depth-dependent scaling rules used to train \modelname{}. Following CompleteP~\citep{dey2025completep}, we apply an output scaling factor of $L^{-1/2}$ to both the attention and FFN branches. For the attention branch, this factor is incorporated into the output RMSNorm. For the FFN branch, it is applied directly to the branch output.

\begin{table*}[t]
\centering
\small
\setlength{\tabcolsep}{4pt}
\begin{tabular}{p{0.22\textwidth}p{0.22\textwidth}p{0.22\textwidth}}
\toprule
\textbf{Module} & \textbf{Initialization Std.} & \textbf{Learning Rate} \\
\midrule
Embedding & $\mathcal{O}(1)$ & $\mathcal{O}(1)$ \\
\midrule
Linear & $\mathcal{O}(d^{-1/2})$ & $\mathcal{O}(d^{-1/2}L^{-1/4})$ \\
RMSNorm & -- & $\mathcal{O}(L^{-1/4})$ \\
\midrule
LM head & $\mathcal{O}(d^{-1})$ & $\mathcal{O}(d^{-1})$ \\
\bottomrule
\end{tabular}
\caption{Width- and depth-dependent $\mu$P scaling rules used for different modules. Here, $d$ denotes the hidden width and $L$ denotes the model depth.}
\label{tab:mup_scaling_factors}
\end{table*}

\begin{figure*}[t]
    \centering
    \begin{subfigure}[t]{0.48\textwidth}
        \centering
        \includegraphics[width=\linewidth]{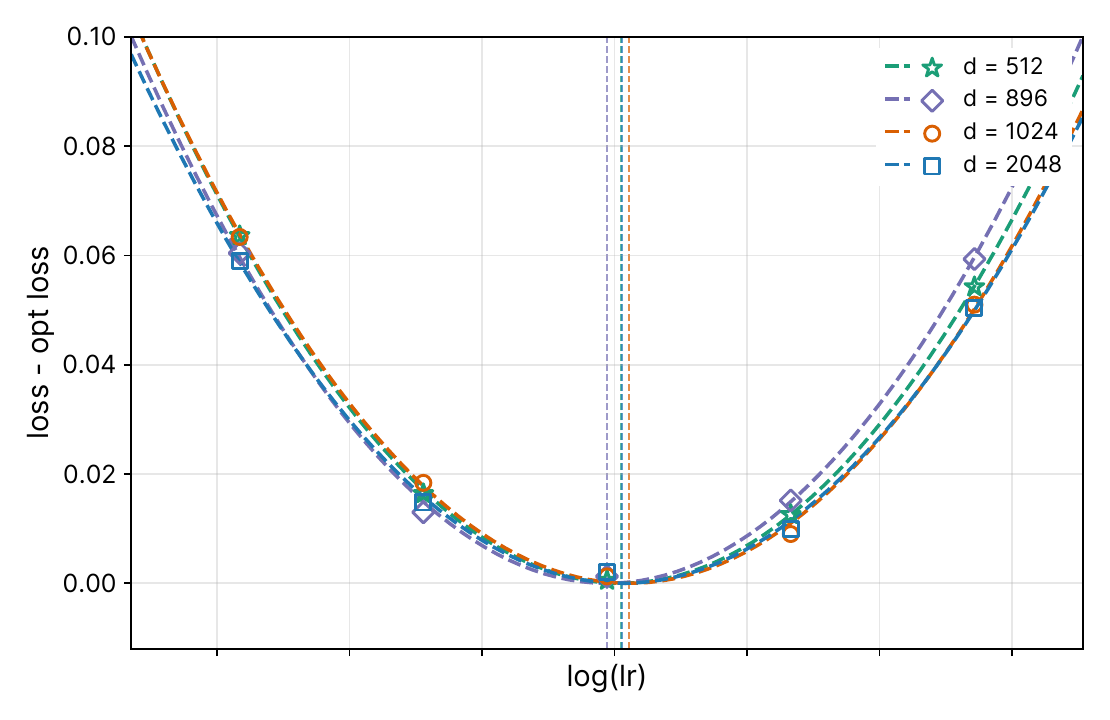}
        \caption{Across different widths.}
        \label{fig:width_mup}
    \end{subfigure}
    \hfill
    \begin{subfigure}[t]{0.48\textwidth}
        \centering
        \includegraphics[width=\linewidth]{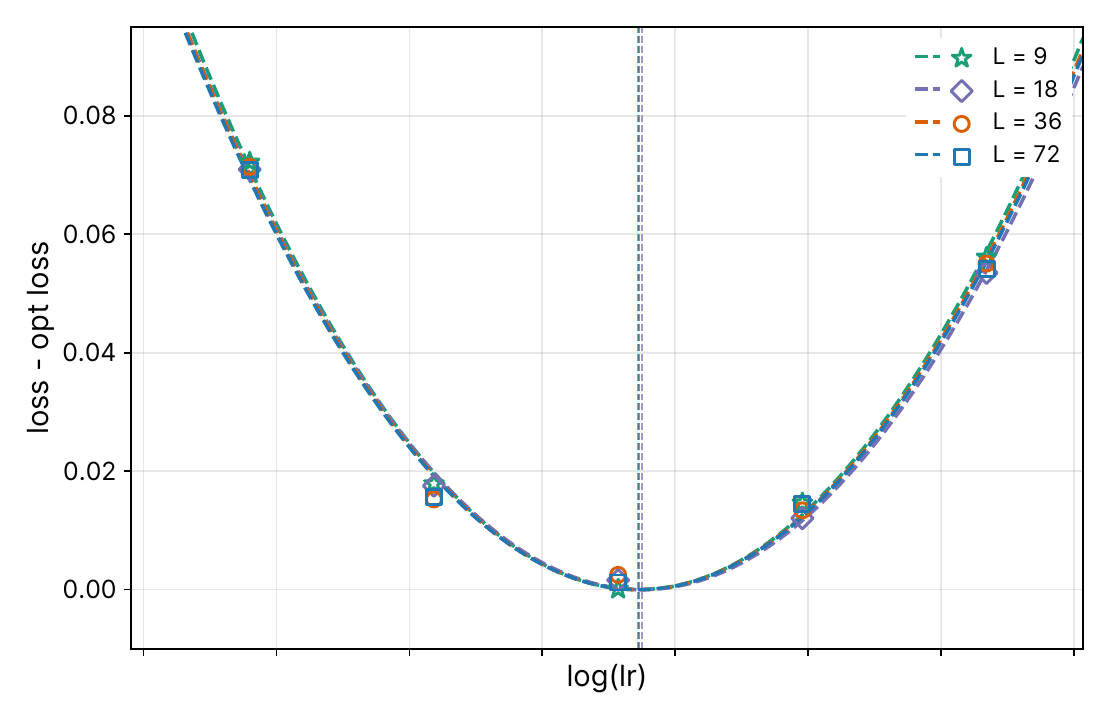}
        \caption{Across different depths.}
        \label{fig:depth_mup}
    \end{subfigure}
    \caption{Optimal learning rate transfer across model widths and depths under $\mu$P scaling.}
    \label{fig:mup}
\end{figure*}

As shown in Figure~\ref{fig:mup}, these scaling rules enable consistent transfer of the optimal learning rate across models with different widths and depths.

\begin{figure}[t]
    \centering
    \includegraphics[width=\linewidth]{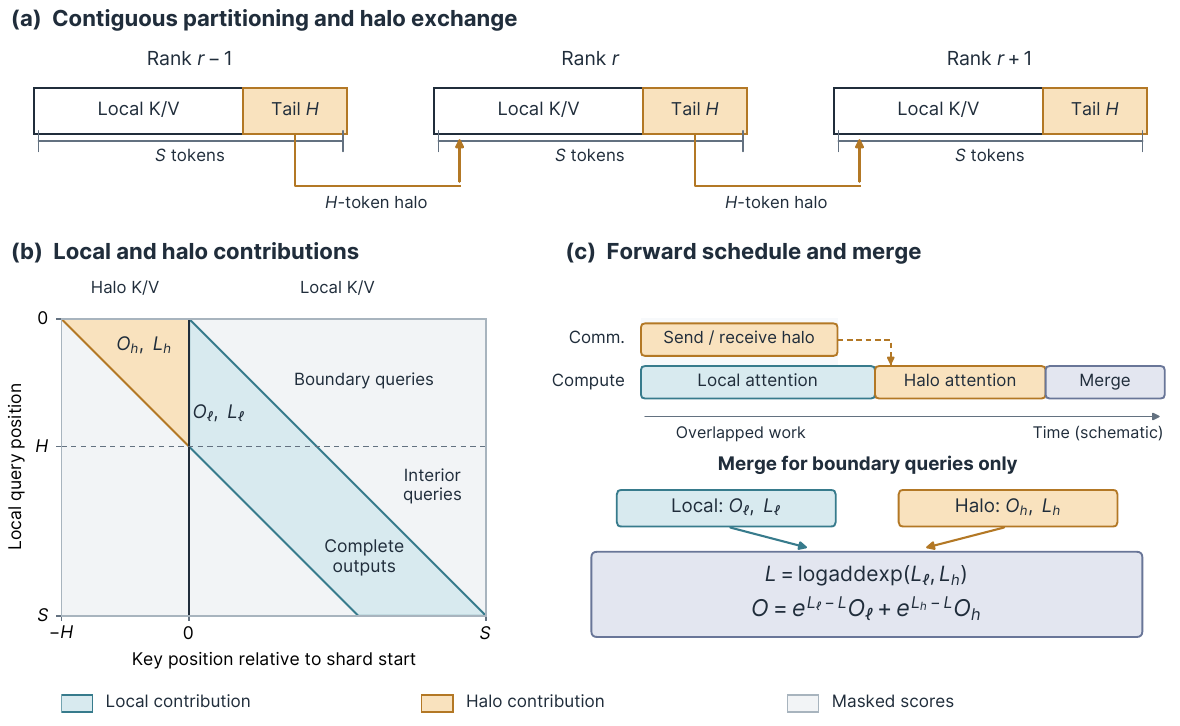}
    \caption{Overlapping halo exchange with partial attention computation.}
    \label{fig:swa_halo_partial_output}
\end{figure}

\paragraph{SWA linear context parallelism with partial halo attention.}
\autoref{fig:swa_halo_partial_output} illustrates the setting $0<H<S$, where $S$ denotes the number of tokens in each contiguous sequence shard and $H$ denotes the number of preceding tokens accessible to each query. In panel~(b), query and key positions are measured relative to the start of the local shard. For a query at position $i$, valid key positions satisfy $i-H \leq j \leq i$; negative key positions belong to the preceding shard. The horizontal dashed line separates boundary queries from queries whose attention windows lie entirely within the local shard. The pairs $(O_\ell,L_\ell)$ and $(O_h,L_h)$ denote the normalized attention outputs and corresponding per-query, per-head log-sum-exp statistics for the local and halo contributions, respectively. Panel~(c) depicts execution dependencies schematically.

\begin{figure}[t]
    \centering
    \includegraphics[width=\linewidth]{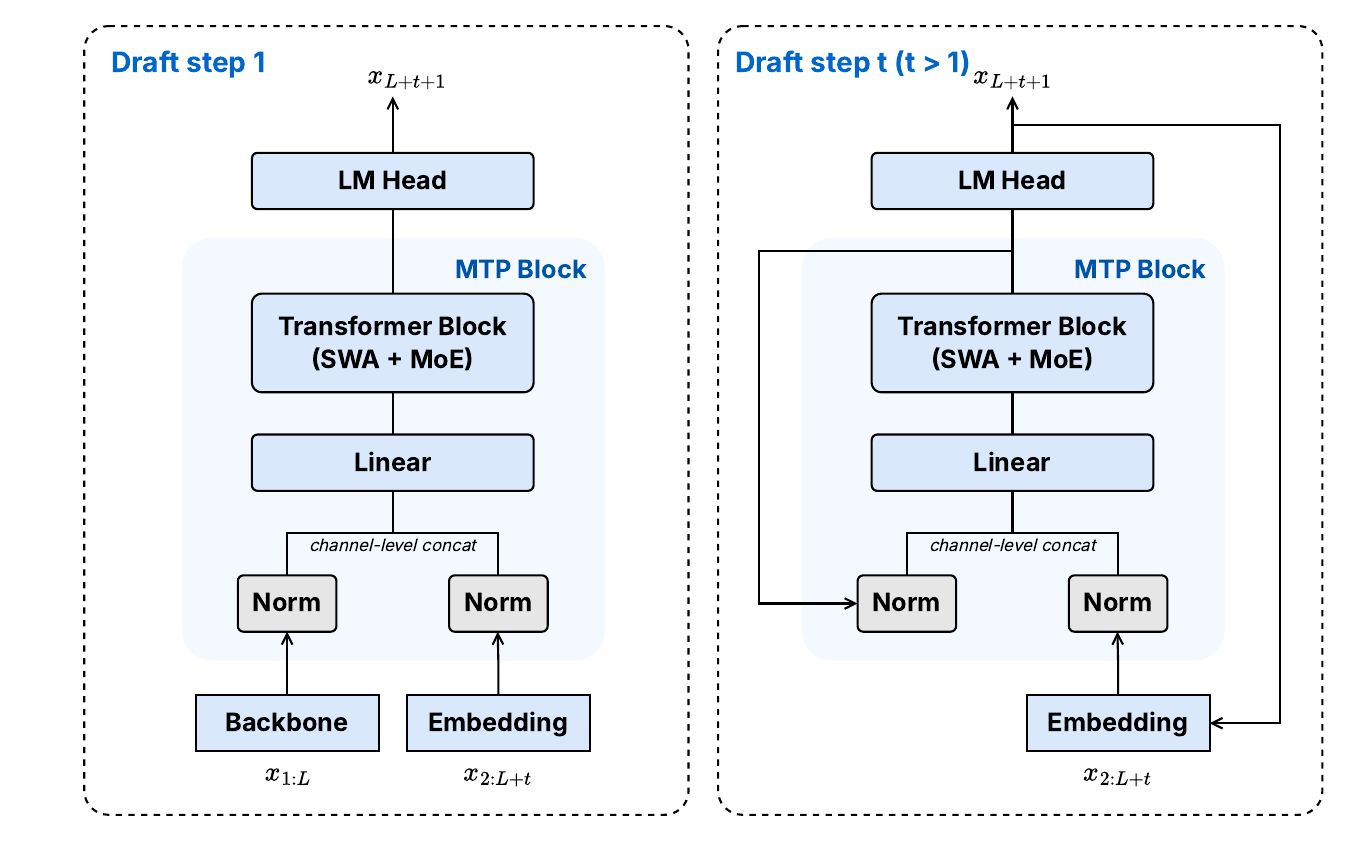}
    \caption{Illustration of recursive multi-token prediction. Here, $L$ denotes the input sequence length and $t$ denotes the drafting step.}
    \label{fig:mtp}
\end{figure}

\paragraph{Recursive Multi-token Prediction.}
\autoref{fig:mtp} illustrates the workflow of recursive multi-token prediction. We adopt the MTP architecture of DeepSeek-V3~\citep{deepseekai2024deepseekv3technicalreport}, which concatenates a hidden representation with a token embedding to form the input to the MTP block. At the first drafting step, the hidden representation comes from the final layer of the backbone model. At each subsequent step, we concatenate the previous step’s output hidden representation with the embedding of the token predicted at that step and feed the result back into the same MTP block, enabling recursive draft generation with shared parameters.

